\documentclass[conference]{IEEEtran}
\IEEEoverridecommandlockouts

\usepackage{cite}
\usepackage[OT1]{fontenc} 
\usepackage{amsmath,amssymb,amsfonts}
\usepackage{algorithmic}
\usepackage{graphicx}
\usepackage{booktabs}
\usepackage{multirow}
\usepackage{textcomp}
\usepackage{xcolor}
\usepackage{comment}
\usepackage[ruled,norelsize]{algorithm2e}

\def\BibTeX{{\rm B\kern-.05em{\sc i\kern-.025em b}\kern-.08em
    T\kern-.1667em\lower.7ex\hbox{E}\kern-.125emX}}

\usepackage{stfloats}

\begin{document}

\title{A Digital Twin Framework for Metamorphic Testing of Autonomous Driving Systems Using Generative Model}
\author{
    \IEEEauthorblockN{Tony Zhang$^{*}$, Burak Kantarci$^{*}$, Umair Siddique$^{**}$}
    \IEEEauthorblockA{
        $^{*}$School of Electrical Engineering and Computer Science, University of Ottawa, Canada\\
        \texttt{\{yzhan117,burak.kantarci\}@uOttawa.ca}\\[2mm]
        $^{**}$reasonX Labs Inc., Ottawa, Canada\\
        \texttt{umair@reasonx.ai}
    }
}

\maketitle

\begin{abstract}
Ensuring the safety of self-driving cars remains a major challenge due to the complexity and unpredictability of real-world driving environments. Traditional testing methods face significant limitations, such as the oracle problem, which makes it difficult to determine whether a system's behavior is correct, and the inability to cover the full range of scenarios an autonomous vehicle may encounter. In this paper, we introduce a digital twin-driven metamorphic testing framework that addresses these challenges by creating a virtual replica of the self-driving system and its operating environment. By combining digital twin technology with AI-based image generative models such as Stable Diffusion, our approach enables the systematic generation of realistic and diverse driving scenes. This includes variations in weather, road topology, and environmental features, all while maintaining the core semantics of the original scenario. The digital twin provides a synchronized simulation environment where changes can be tested in a controlled and repeatable manner. Within this environment, we define three metamorphic relations inspired by real-world traffic rules and vehicle behavior. We validate our framework in the Udacity self-driving simulator and demonstrate that it significantly enhances test coverage and effectiveness. Our method achieves the highest true positive rate (0.719), F1 score (0.689), and precision (0.662) compared to baseline approaches.
This paper highlights the value of integrating digital twins with AI-powered scenario generation to create a scalable, automated, and high-fidelity testing solution for autonomous vehicle safety.
\end{abstract}


\section{Introduction}
The complexity of real-world situations makes the development and certification of autonomous driving systems (ADS) \cite{zhao2023autonomous} extremely difficult.  ADS must function in dynamic, unexpected contexts that necessitate the use of complex testing procedures, in contrast to standard software systems with clearly defined inputs and outputs. This issue becomes even more challenging due to the fact that many modern ADS architectures operate as black-box systems \cite{9284628}, which conceals their decision-making processes and makes it difficult to comprehensively validate their outcomes.

A possible method for verifying ADS, especially in situations without a conclusive test oracle, is Metamorphic Testing (MT)\cite{chen2018metamorphic} in combination with a Digital Twin\cite{mihai2022digital}. By analyzing invariant relations between outputs when inputs go through regulated transformations, MT assesses the behavior of the system\cite{zhang2025augmented} and Digital Twin provides a dependable platform for consistent and repeatable testing\cite{9628341}. However, traditional MT methods frequently depend on simple transformations and restricted metamorphic connections \cite{segura2016survey}.This simplicity can lead to systems that adapt to specific test patterns rather than developing genuine robustness, potentially resulting in models that perform adequately under test conditions but fail to generalize to real-world scenarios \cite{10.1145/3597503.3639191}.

\begin{figure}[h!]
    \centering
    \includegraphics[width=0.95\columnwidth]{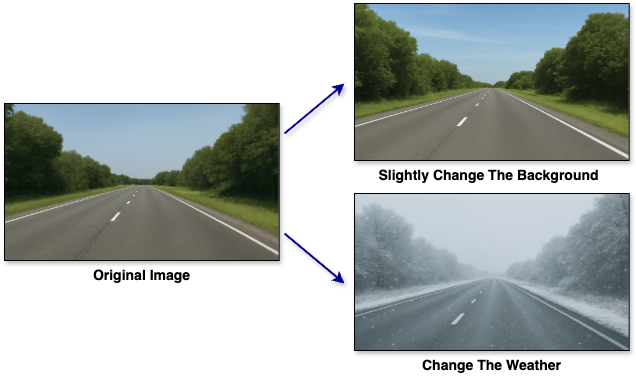}
    \caption{The case of using a generative model to apply transformations to a real image.}
\end{figure}

To address these limitations, we propose incorporating advanced generative models, particularly Stable Diffusion \cite{rombach2021highresolution}. While preserving environmental coherence, these generative models are excellent at creating Digital Twin  and generating complex, controlled variations from input data \cite{10419041}. We generate complex scenario variations—such as changes in lighting, weather, and lane layout—within the MT framework to fully assess ADS decisions. Using the Stable Diffusion-XL model, we subtly alter backgrounds while preserving the original lane direction in Figure 1 , creating diverse metamorphic test cases.

Our methodology creates reusable Digital Twins based on ADS behavioral invariants and generates consistent but varied scenarios to increase test reproducibility. Path planning and obstacle detection are two important ADS components that we assess using generative transformations to create a Digital Twin on camera inputs\cite{8627998}. By overcoming the limitations of traditional MT and leveraging modern generative models, our framework achieves an F1 score of 0.689 and a precision of 0.662 significantly outperforming baseline methods such as CAE and SAE.

\section{Background}

To ensure ADS safety, prior work has applied metamorphic testing and uncertainty-aware methods to detect anomalies in rare or uncertain scenarios.
Digital twin frameworks have gained traction for enhancing runtime safety and anomaly detection in autonomous driving systems (ADS). Kumar et al. (2025) introduced the use of digital twins to assess the feasibility of operational failure detection and decision-making in autonomous vehicles\cite{11086226}. Similarly, the AutoDRIVE ecosystem by Samak et al. (2023) offers a modular, flexible, and open-access digital twin platform that enables hardware–software co-simulation, real-time testing, and scalable validation for both research and educational purposes in autonomous driving\cite{samak2023autodrive}.  Wu et al. (2021) proposed a digital twin-enabled reinforcement learning framework that integrates a predictive environment model and rollout-compatible training to improve data efficiency and performance in autonomous driving control tasks\cite{9540179}.
Stocco et al. (2020) introduced SelfOracle, a runtime monitoring technique that estimates DNN confidence using autoencoders and time-series anomaly detection to forecast potential misbehaviors, achieving a 77\% detection rate \cite{stocco2020selforacle}. However, it lacks domain-specific metamorphic reasoning and may not generalize to unseen scenarios. Wang et al. (2023) proposed uncertainty-aware metamorphic testing for object detection, using Bayesian uncertainty to evaluate robustness under realistic perturbations, though it remains limited to offline use \cite{wang2023uncertainty}. Most recently, Ayerdi et al. (2024) developed MarMot, a runtime monitoring approach that leverages domain-specific metamorphic relations to detect both external and internal anomalies in ADSs, outperforming prior methods like SelfOracle and MC Dropout in both simulated and real-world settings\cite{ayerdi2024marmot}.

Prior studies improve DNN-based ADS robustness with metamorphic or uncertainty-aware methods, but fixed transformations (e.g., brightness, rotation) are easy for models to ignore and miss complex cases like agent interactions or road topology changes. With digital twin–supported generative models (e.g., Stable Diffusion), we can create diverse, realistic driving scenes, enabling richer transformations for more effective ADS testing under challenging conditions.

The validation of Autonomous Driving Systems (ADS) necessitates a comprehensive approach that accounts for both methodological rigor and real-world operational limitations. This section discusses three interrelated concepts critical to this validation process: the Operational Design Domain (ODD) \cite{ISO34503:2023}, Metamorphic Testing (MT), and the application of generative models \cite{oussidi2018deep}.

\subsection{Operational Design Domain}
ODD specifies conditions for ADS functionality, defined as a tuple: $P$ (infrastructure), $E$ (environment), $O$ (constraints), $T$ (temporal), and $C$ (connectivity) , as shown in Eq.\ref{eq:ODD}.

\begin{equation}
    \label{eq:ODD}
    \text{ODD} = (P, E, O, T, C)
\end{equation}

To capture the complexity of environmental factors, each component particularly the $E$(environment) can be broken down into the following parameters\cite{10186765}, as defined in Eq.\ref{eq:Ecomp}.

\begin{equation}
    \begin{split}
        E = \{&e_w \text{ (weather)}, e_l \text{ (lighting)}, \\
        &e_v \text{ (visibility)}, e_t \text{ (temperature)}\}
    \end{split}
    \label{eq:Ecomp}
\end{equation}

\subsection{Metamorphic Testing with ODD}
Metamorphic Testing in a specified ODD involves the conditions that the metamorphic relations\cite{inbook} should hold within certain operational limits. For any ADS $S$, input domain $I$, and output domain $O$, the ODD-constrained metamorphic relations are defined as in Eq.\ref{eq:MRODD}.
\begin{equation}
    \begin{split}
        \text{MR}_{\text{ODD}} \subseteq \{&(x, S(x), x', S(x')) \mid \\
        &x, x' \in I_{\text{ODD}}, \\
        &R(x, S(x), x', S(x')) = \text{true}\}
    \end{split}
    \label{eq:MRODD}
\end{equation}

where $I_{\text{ODD}}$ represents inputs valid within the ODD constraints, as defined in Eq.\ref{eq:I_ODD}.

\begin{equation}
    I_{\text{ODD}} = \{x \in I \mid \forall c \in \text{ODD}: V(x,c) = \text{true}\}
    \label{eq:I_ODD}
\end{equation}

Here, $V(x,c)$ verifies compliance with ODD constraint $c$.

\subsection{Generative Models with ODD Integration}
We extend the generative model framework to incorporate ODD constraints. For a generative model $G$ and manually defined transformation specification $\tau$ , as shown in Eq. \ref{eq:G_ODD}.

\begin{equation}
    \label{eq:G_ODD}
    G(x, \tau, \text{ODD}) \rightarrow x' \text{ where } x, x' \in I_{\text{ODD}}
\end{equation}

The transformation specification $\tau$ is now ODD-aware, as formally defined in Eq. \ref{eq:tau_ODD}:

\begin{equation}
\label{eq:tau_ODD}
\tau_{\text{ODD}} = {\varepsilon \in E, \gamma \in P, \sigma \in O \times T \times C}
\end{equation}

This enables the definition of ODD-compliant metamorphic relations as shown in Eq. \ref{eq:MRrel}:

\begin{equation}\label{eq:MRrel} 
    \begin{split}
        \text{MR}_{G,\text{ODD}} = \{&(x, S(x), G(x,\tau_{\text{ODD}}), \\
        &S(G(x,\tau_{\text{ODD}}))) \mid \\
        &R(x, S(x), G(x,\tau_{\text{ODD}}), \\
        &S(G(x,\tau_{\text{ODD}}))) = \text{true} \wedge \\
        &x, G(x,\tau_{\text{ODD}}) \in I_{\text{ODD}}\}
    \end{split}
\end{equation}

The formulation ensures synthesized cases remain within ODD limits, retain key ODD characteristics, and produce valid, meaningful results within the intended context.

Integrating ODD with MT and generative models to ensure that ADS testing remains relevant within the intended conditions, while also upholding the rigor of mathematical validation\cite{10591490}.

\section{Proposed Approach}

ADSs demand thorough validation within their ODD. We propose a novel framework that integrates MT with generative AI to systematically validate ADS perception systems, addressing three key challenges: 1) Oracle problem \cite{barr2014oracle} in ADS testing, 2) environmental complexity and scenario diversity, 3) uncertainty in perception systems.
\begin{algorithm}[h]
\caption{ODD-Aware Digital Twin Scenario Generation}
\SetAlgoLined
\KwIn{Source image $x$, ODD specifications}
\KwOut{Transformed image $x'$}
Define $\tau$ based on ODD constraints\;
Verify transformation validity: $V(x,\tau) = \text{true}$\;
Generate candidate: $x' \leftarrow G_{\text{ODD}}(x,\tau)$\;
\eIf{ValidateODDCompliance($x'$)}{
    \Return{$x'$}\;
}{
    \Return{GenerateNewTransform($x$)}\;
}
\end{algorithm}

\begin{figure*}[h]
    \centering
    \includegraphics [width=0.95\textwidth]{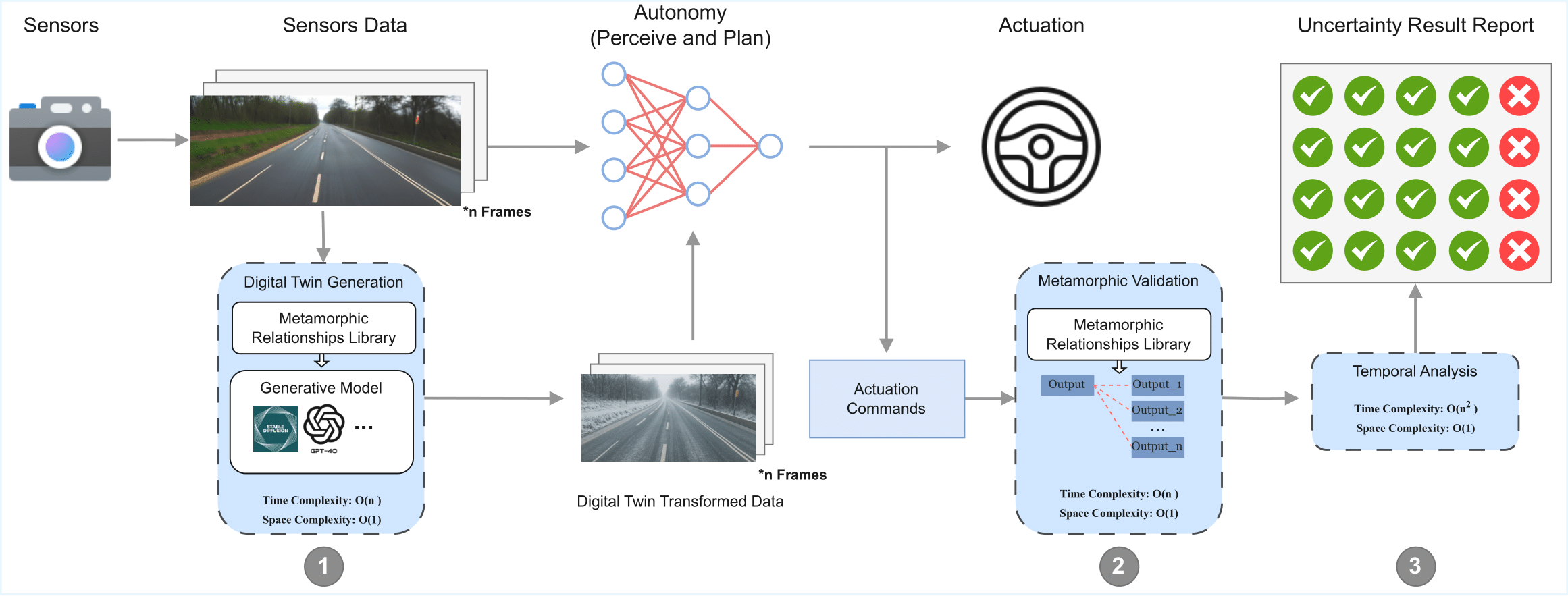} 
    \caption{Architecture of Digital Twin for metamorphic testing of ADS and its three key components: (1) Digital Twin Generation, (2) Integrated Validation, (3) Time Series Analysis.}
    \label{fig:arch}
\end{figure*}

\subsection{Framework Overview}
Our framework operates through a systematic workflow (Fig. 2) with two core components:
\begin{itemize}
    \item \textbf{Digital Twin Scenario Generation}: Creates controlled variations in test scenarios while maintaining ODD compliance.
    \item \textbf{Metamorphic Validation}: Evaluates ADS behavior consistency under these variations.
    \item \textbf{Temporal Analysis}: Ensures consistency of predictions over time using smoothed outputs.
\end{itemize}

Formally, let $S$ represent an ADS under test, as defined in Eq \ref{eq:ADS_model}, where $I$ denotes the input space (camera images), $O$ represents the output space (driving decisions), and $h$, $w$, and $c$ stand for image height, width, and channels, respectively:

\begin{equation}
\label{eq:ADS_model}
S: I \rightarrow O, \text{ where } I \subset \mathbb{R}^{h \times w \times c}
\end{equation}

The framework components are defined in Eq \ref{eq:framework_components}:

\begin{equation}
    \label{eq:framework_components}
    \begin{split}
        G_{\text{ODD}}&: I \times \tau \rightarrow I \\
        V_{\text{MR}}&: (I \times O) \times (I \times O) \rightarrow \{0,1\}
    \end{split}
\end{equation}

where $G_{\text{ODD}}$ generates ODD-compliant transformations and $V_{\text{MR}}$ validates metamorphic relations.

\subsection{ODD-Aware Digital Twin Scenario Generation}

\subsubsection{Transformation Space}
For a source image $x \in I$, we define ODD-compliant transformations with the following components:
     Environmental conditions  such as weather and lighting ($\varepsilon$),
    geometric transformations such as perspective and scale ($\gamma$), and
  semantic modifications such as objects, road features ($\sigma$) as shown in Eq. \ref{eq:TODDsubject}.

\begin{equation}
    \label{eq:TODDsubject}
    \begin{split}
        \tau_{\text{ODD}} = \{&\varepsilon \in E, \gamma \in P, \sigma \in O\} \\
        \text{subject to: } &\forall c \in \text{ODD}: V(G(x,\tau),c) = \text{true}
    \end{split}
\end{equation}

\subsubsection{Generation Process}
involves generation of transformed images that adhere to ODD specifications, utilizing a visual generation model to produce metamorphic testing samples. This is detailed in Algorithm 1, and also illustrated as an integral component (1) of the metamorphic testing architecture for ADS in Fig. \ref{fig:arch}.

\subsection{Metamorphic Relations and Validation}

\subsubsection{Uncertainty-Aware Relations}
We enhance traditional MRs with uncertainty quantification as follows where $u(\cdot)$ denotes uncertainty quantification, $\theta_u$ represents the uncertainty threshold, and $R(\cdot,\cdot)$ stands for the relation validator as shown in Eq. \ref{eq:MRu}:

\begin{equation}
    \label{eq:MRu}
    \begin{split}
        \text{MR}_u(x,x') = \{&(S(x), S(x'), u(S(x)), u(S(x'))) \mid \\
        &R(S(x), S(x')) = \text{true} \wedge \\
        &u(S(x')) \leq \theta_u\}
    \end{split}
\end{equation}

\subsubsection{Validation Criteria}
For each MR category validation criteria are formulated as follows with the following three key components: 
    Path extraction ($P(\cdot)$), Object detection ($D(\cdot)$), and tolerance thresholds ($\epsilon_p, \epsilon_d$).

MR1 , MR2 and MR3 require that the error threshold is not exceeded, as the generator utilizes similar images for testing. The metamorphic relations are defined as Eq. \ref{eq:MRDefine}

\begin{equation}
    \label{eq:MRDefine}
    \begin{split}
        V_{\text{MR1,2,3}}(x,x') &= \|P(S(x)) - P(S(x'))\| \leq \epsilon_p \\
        V_{\text{MR1,2,3}}(x,x') &= \|D(S(x)) - D(S(x'))\| \leq \epsilon_d \\
    \end{split}
\end{equation}

\begin{itemize}
    \item \textbf{MR1}: Maintain the same lane direction and angle, while slightly altering the background. The expectation is that the result of the test case should remain within an error margin of $\epsilon_p $, since the key lane features are preserved.

    \item \textbf{MR2}: Keep the lane's direction and angle unchanged, but modify the weather conditions to snow. Snow partially obscures the road view, simulating a realistic environmental effect on perception algorithms. Despite the occlusion, the output should remain consistent with the original, within an acceptable error margin $\epsilon_d $.

    \item \textbf{MR3}: Maintain the same lane direction and angle while narrowing the driving lane. The expected outcome of the test case should remain within an error margin of $\epsilon_e $, as the primary direction of the lane features is preserved.

\end{itemize}

\subsection{Temporal Analysis}

The third key component of the metamorphic testing framework in Fig. \ref{fig:arch} is time series analysis which aims to ensure robust validation across time sequences as formulated below where   $w$, $\epsilon_t$, and $S_t(\cdot)$ denote time window size, temporal threshold and smoothed prediction, respectively. The temporal validation condition is given in Eq \ref{eq:St}.

\begin{equation}
    \label{eq:St}
    \begin{split}
        S_t(x) &= \frac{1}{w}\sum_{i=t-w}^t S(x_i) \\
        V_{\text{temporal}}(x,x') &= \|S_t(x) - S_t(x')\| \leq \epsilon_t
    \end{split}
\end{equation}

\subsection{Integrated Validation Framework}
Integrated validation evaluates if outputs meet safety standards by utilizing metamorphic relations, as mathematically defined and described in Algorithm 2. This framework acts as the second essential part of the metamorphic testing architecture for ADS, as depicted in Fig. \ref{fig:arch}

\begin{algorithm}[h]
\caption{Integrated Validation Framework}
\SetAlgoLined
\KwIn{Image sequence $X$, ADS $S$, ODD specifications}
\KwOut{Validation report $R$}
Initialize empty report $R$\;
\ForEach{$x_t \in X$}{
    Generate ODD-compliant $\tau_t$\;
    $x'_t \leftarrow G_{\text{ODD}}(x_t, \tau_t)$\;
    $s_t \leftarrow S(x_t)$\;
    $s'_t \leftarrow S(x'_t)$\;
    $u_t \leftarrow$ ComputeUncertainty($s_t$)\;
    $u'_t \leftarrow$ ComputeUncertainty($s'_t$)\;
    $v_{\text{mr}} \leftarrow$ ValidateRelations($s_t, s'_t$)\;
    $v_{\text{temp}} \leftarrow$ TemporalValidation($s_{t-w:t}, s'_{t-w:t}$)\;
    UpdateReport($R, v_{\text{mr}}, v_{\text{temp}}, u_t ,u'_t $)\;
}
\Return{$R$}\;
\end{algorithm}

This systematic framework enables rigorous validation of ADS perception systems while maintaining practical relevance within specified operational bounds.


\section{Empirical Evaluation}
\subsection{Procedure}
We generated a dataset of transformed driving scenarios using Stable Diffusion-XL as described in Section III. For each base driving frames within the dataset, we applied controlled variations to simulate diverse ODD conditions such as fog, rain, glare, and darkness. Each ODD-compliant transformation was validated using Algorithm 1, ensuring operational integrity.
\subsection{Setups}

\begin{figure}[h!]
    \centering
    \includegraphics[width=0.48\textwidth]{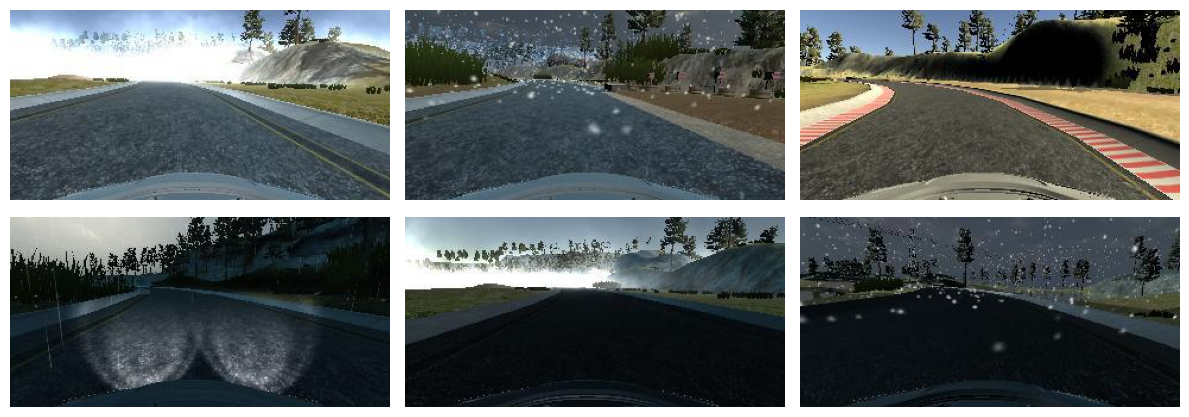} 
    \caption{Sample scenes from test dataset, which includes diverse driving scenarios across different times of day and weather conditions.}
\end{figure}

To unify the measurement standards, we use the dataset that contains crash cases collected in the Udacity simulator under various weather and conditions, which is the same as that used in the Self Oracle project\cite{stocco2020selforacle}. These episodes encompass a wide range of environmental conditions and temporal settings, including variations such as Daynight, Daynightfog, Daynightrain, and Daynightsnow, as well as isolated conditions like Fog, Rain, Snow, and Normal weather. This comprehensive coverage enables robust evaluation of the model's performance under realistic and challenging circumstances.

For our test model, we adopted the DAVE-2 architecture with nine layers: one normalization, five convolutional, and three fully connected layers. Input images were split into YUV planes and the model was trained on the Udacity Simulator Dataset across three tracks with dropout rates between $0.1$ and $0.5$. For metamorphic test generation, we used Stable Diffusion XL with $16$-bit precision, processing $1024 \times 512$ pixel images with transformation strength $0.2$, guidance scale $10.0$, and negative prompts excluding ``low quality, distorted, cartoonish, unrealistic'' content.

\subsection{Results}

\begin{table}[!h]
\centering
\resizebox{\columnwidth}{!}{%
\begin{tabular}{@{}lcccccccc@{}}
\toprule
\textbf{Method} & \textbf{TP} & \textbf{FP} & \textbf{TN} & \textbf{FN} & \textbf{TPR} & \textbf{FPR} & \textbf{F1} & \textbf{Prec.} \\
\midrule
\multicolumn{9}{l}{\textbf{\textit{SelfOracle}}} \\
\quad CAE & 0 & 0 & 4282 & 196 & 0 & 0 & n.a. & n.a. \\
\quad DAE & 21 & 55 & 4063 & 175 & 0.107 & 0.013 & 0.154 & 0.276 \\
\quad SAE & 108 & 183 & 2665 & 88 & 0.551 & 0.064 & 0.444 & 0.371 \\
\quad VAE & 107 & 183 & 2959 & 89 & 0.546 & 0.058 & 0.440 & 0.369 \\
\textbf{LSTM} & 7 & 12 & 4119 & 186 & 0.036 & \textbf{0.003} & 0.066 & 0.368 \\
\textbf{DeepRoad} & 44 & 250 & 3651 & 152 & 0.225 & 0.064 & 0.180 & 0.150 \\
\multicolumn{9}{l}{\textbf{\textit{Stable Diffusion}}} \\
\quad MR1 & 137 & 77 & 403 & 59 & 0.699 & 0.160 & 0.668 & 0.640 \\
\quad MR2 & 141 & 72 & 408 & 55 & \textbf{0.719} & 0.150 & \textbf{0.689} & \textbf{0.662} \\
\quad MR3 & 135 & 87 & 393 & 61 & 0.689 & 0.181 & 0.645 & 0.608 \\
\bottomrule
\end{tabular}%
}
\vspace{2pt}
\caption{Evaluation results for all variants.}
\label{tab:evaluation_results_grouped}
\end{table}
Compared to Self-Oracle and DeepRoad, our Stable Diffusion variants (MR1,MR2,MR3) show notable gains in TPR, F1, and Precision—key metrics for reliable, accurate prediction in safety-critical applications.

Among the evaluated methods: MR2 consistently outperforms all other strategies, achieving the highest TPR (0.719), F1 score (0.689), and Precision (0.662). This suggests that MR2 is not only more accurate in detecting true crash scenarios but also less prone to false alarms compared to other methods.

MR1 and MR3 also show robust performance, with F1 scores of 0.668 and 0.645, respectively. While MR3 shows a slightly higher TPR than MR1 (0.689 vs. 0.699), its FPR is also marginally higher, indicating a small trade off between sensitivity and specificity. In contrast, traditional Self-Oracle approaches such as VAE, DAE, SAE, and DeepRoad fall behind. For instance, DeepRoad has a TPR of only 0.225 and an F1 score of 0.180, reflecting weaker detection capabilities.
These improvements are further illustrated by Figure 4, which tracks the number of successful crash predictions over time for each MR variant. The plot illustrates that:

MR3 highlights the most favorable early crash prediction performance, with a distribution that is more concentrated in earlier frames, particularly within the critical 5-second pre-crash window(represented by the dashed blue line). Although MR2 achieves the highest total number of successful predictions (141 out of 196), its predictions are more concentrated closer to the crash event, similar to MR1. MR1 and MR2 exhibit comparable patterns in both timing and peak predictive success, though they are slightly less proactive than MR3 in anticipating crashes.

\begin{figure}[!h]
    \centering
    \includegraphics[width=0.48\textwidth]{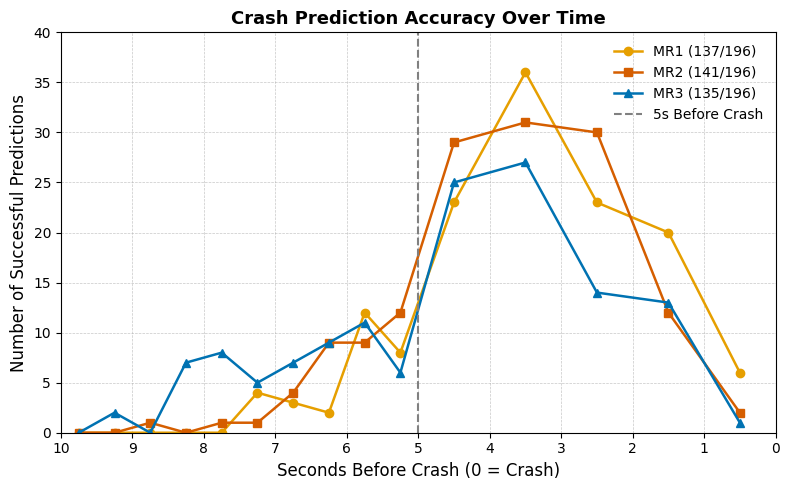} 
    \caption{Distribution of successful crash predictions made by each MR variant across time leading up to the crash.}
\end{figure}

The time series trend highlights the proposed method’s ability not only to respond to imminent crashes but also to act as a proactive safety mechanism, detecting potential hazards well before they occur.

These findings demonstrate the method’s superior effectiveness, strong early warning capability, and consistent reliability compared to traditional approaches. They also confirm its suitability for real-world deployment in autonomous systems and other safety-critical settings, where timely and accurate hazard prediction is vital.

\subsection{Extensions and Challenges of the Proposed Framework}

By integrating a generative model, our framework enables the design of flexible and adaptive metamorphic relations beyond the three initially proposed. Due to the use of the generative model and the digital twin feature, our framework offers high flexibility in creating a variety of metamorphic rules. Owing to content limitations, we explored the three most representative and practical metamorphic relations in this paper. As shown in Table~\ref{tab:more-mr-table}, we list the potential for the proposed framework to generate other efficient test cases.

Although the framework shows exceptional performance, the significant processing power demands of Stable Diffusion models create computational hurdles for real-time implementation. Consequently, the approach is presently best suited for closed-loop testing settings during the development and certification stages of ADS. With advancements in generative model technology and improvements in computational efficiency via methods like model compression and specialized hardware acceleration, the framework's use will broaden to include real-time tracking and production implementation in autonomous vehicles

\begin{table*}[!h]
\caption{Potential Metamorphic Relations for Autonomous Driving Scenarios}
\begin{center}
\begin{tabular}{|c|p{3cm}|p{6cm}|p{5.5cm}|}
\hline
\textbf{MR} & \textbf{Name} & \textbf{Transformation ($\tau$)} & \textbf{Description} \\
\hline
MR4 & Agent Substitution & $\tau_{\text{agent}}(x) =$ GenerateAgent(x, new\_type, preserve=(position, velocity, size\_class)) & Replace traffic participants with similar-sized agents (car$\leftrightarrow$truck, bike$\leftrightarrow$motorcycle) while preserving position and velocity \\
\hline
MR5 & Time-of-Day Consistency & $\tau_{\text{time}}(x) =$ AdjustLighting(x, target\_time, preserve=(geometry, objects, lanes)) & Transform day scenes to night/dusk while preserving all geometric and semantic content \\
\hline
MR6 & Traffic Control Equivalence & $\tau_{\text{control}}(x) =$ ReplaceSignal(x, equivalent\_control, preserve=(intersection\_geometry)) & Replace equivalent traffic control devices (stop sign $\leftrightarrow$ red light) \\
\hline
MR7 & Emergency Vehicle Priority & $\tau_{\text{emergency}}(x) =$ ReplaceEmergency(x, new\_type, preserve=(priority, position, signals)) & Replace emergency vehicle type while maintaining priority status and position \\
\hline
MR8 & Construction Zone Adaptation & $\tau_{\text{construction}}(x) =$ AddConstruction(x, cone\_pattern, preserve=(intended\_path, lane\_width)) & Transform normal lanes to construction zones with equivalent path guidance \\

\hline
MR10 & Obstacle Substitution & $\tau_{\text{obstacle}}(x) =$ ReplaceObstacle(x, equivalent\_obstacle, preserve=(size, position, blockage)) & Replace static obstacles with equivalent objects (fallen tree$\rightarrow$barrier) \\
\hline
\end{tabular}
\end{center}

\label{tab:more-mr-table}
\end{table*}

\section{Conclusion}
This paper presents a Digital Twin based approach to advancing the validation of autonomous driving systems (ADS) by enabling systematic evaluation of system behavior across a wide range of driving scenarios, including rare and safety-critical edge cases. By creating a virtual replica of the ADS and its operational environment, the framework facilitates controlled, repeatable testing through reusable metamorphic relations and robust evaluation metrics. This digital twin integration enhances safety assurance and supports the development of resilient machine learning components for real-world deployment. Through experiments conducted in the Udacity Simulator, we implemented and evaluated the framework against baseline methods such as Self-Oracle and DeepRoad. Among the proposed metamorphic relations, MR2 yielded the highest true positive rate (0.719), F1 score (0.689), and precision (0.662), while MR3 exhibited strong early crash prediction performance, with accurate detections concentrated before the critical 5-second mark. These findings highlight the effectiveness of our digital twin-driven method in identifying hazardous behaviors with improved accuracy and responsiveness. Overall, we demonstrate that incorporating digital twins into metamorphic testing provides a scalable and high-fidelity strategy for safety verification in autonomous vehicles. As ADS technologies evolve, the framework's virtual mirroring capability is expected to enable even more proactive, efficient, and context-aware safety assessments in future mobility.

\section*{Acknowledgement}
This work is supported in part by MITACS Accelerate Program under project IT40981, in part by the NSERC CREATE TRAVERSAL program, and in part by Ontario Research Fund-Research Excellence under grant number RE 12-026.
\bibliographystyle{ieeetr}

\vspace{12pt}

\end{document}